# Cross-Species Transfer Learning in Agricultural AI: Evaluating ZebraPose Adaptation for Dairy Cattle Pose Estimation

**Mackenzie Tapp [1], Sibi Chakravarthy Parivendan [1], Kashfia Sailunaz [1] and Suresh Neethirajan [1,2,*]**

[1] Faculty of Computer Science, Dalhousie University, 6050 University Avenue, Halifax, NS B3H 4R2, Canada.
[2] Faculty of Agriculture, Agricultural Campus, P.O. Box 550, Truro, NS B2N 5E3, Canada.
**\*** Correspondence: sneethir@gmail.com

**Abstract:** Pose estimation serves as a cornerstone of computer vision for understanding animal posture, behavior, and welfare. Yet, agricultural applications remain constrained by the scarcity of large, annotated datasets for livestock, especially dairy cattle. This study evaluates the potential and limitations of cross-species transfer learning by adapting ZebraPose - a vision transformer-based model trained on synthetic zebra imagery - for 27-keypoint detection in dairy cows under real barn conditions. Using three configurations - a custom on-farm dataset (375 images, Sussex, New Brunswick, Canada), a subset of the APT-36K benchmark dataset, and their combination—we systematically assessed model accuracy and generalization across environments. While the combined model achieved promising performance (AP = 0.86, AR = 0.87, PCK@0.5 = 0.869) on in-distribution data, substantial generalization failures occurred when applied to unseen barns and cow populations. These findings expose the synthetic-to-real domain gap as a major obstacle to agricultural AI deployment and emphasize that morphological similarity between species is insufficient for cross-domain transfer. The study provides practical insights into dataset diversity, environmental variability, and computational constraints that influence real-world deployment of livestock monitoring systems. We conclude with a call for agriculture-first AI design, prioritizing farm-level realism, cross-environment robustness, and open benchmark datasets to advance trustworthy and scalable animal-centric technologies.

## 1. Introduction

Pose estimation is a foundational task in computer vision that enables the structured analysis of animal or human body regions and their spatial dynamics in images or videos [1, 2]. By localizing anatomical landmarks, pose estimation facilitates automated behavioral interpretation across species and applications. In animals, keypoint or pose detection refers to the automatic recognition and localization of skeletal features such as the head, neck, tail, and limbs from visual data [3]. These keypoints support non-invasive monitoring of behavior and welfare through posture analysis, movement tracking, feeding and drinking behavior, social interaction, growth assessment, and pain or distress detection. For dairy cattle, pose estimation has been applied in estrus detection, lameness diagnosis, and social network analysis, offering major potential for precision livestock farming [4].

Pose detection approaches generally follow top-down, bottom-up, or one-stage strategies [5]. In the top-down approach, a bounding box first isolates the cow, and the model then predicts keypoints such as eyes, limbs, and ears to reconstruct body posture. The bottom-up method reverses this process by detecting keypoints first and grouping them into individual animals, while one-stage models infer objects and keypoints simultaneously. Deep learning architectures such as HRNet, ViTPose, HRFormer, and SwinTransformer have advanced top-down detection; CID and DEKR have enhanced bottom-up inference; and unified frameworks like ED-Pose and CowK-Net represent emerging one-stage approaches [5–7]. Open-source toolkits such as DeepLabCut [8, 9], T-LEAP [10, 11], SLEAP, and DeepPoseKit [12, 13] have accelerated adoption of animal pose estimation in neuroscience and ethology. Yet, progress in livestock research remains hindered by the need for extensive manual annotation, as each frame must include numerous labeled keypoints [14].

A persistent frontier in agricultural AI is cross-species pose detection, which probes the transferability of learned representations across morphologically related species [15]. This study systematically investigates that frontier using ZebraPose [16, 17]—a vision transformer-based pose detection framework originally trained on synthetic zebra imagery—as a test case for cross-species adaptation in dairy cattle monitoring. Rather than assessing only adaptation potential, we critically analyze the limitations and failure modes of transferring a synthetic, zebra-trained model to real-world barn environments. Our experiments combine a custom on-farm dataset with a subset of the benchmark APT-36K dataset [18] containing only cow images, enabling controlled evaluation of both dataset-specific and environmental generalization effects.

We examined three configurations of pre-trained ZebraPose models-trained separately on our dataset, on APT-36K, and on their combination—to quantify transfer-learning barriers in 27-keypoint estimation. Through systematic testing, we identify the key constraints that limit cross-species generalization in agricultural contexts and provide actionable insights for building robust, domain-specific AI systems.

Our main contributions are as follows:
• Systematic evaluation of cross-species pose-estimation limitations and failure modes for agricultural AI deployment;
• Critical analysis of transfer-learning challenges when adapting synthetic-trained models to real barn environments under variable conditions; and
• Comprehensive assessment of generalization barriers across datasets, cow populations, and environmental contexts, establishing realistic performance baselines for future agricultural computer-vision research.

**2. ZebraPose in Pose Estimation**
ZebraPose - originally developed for zebra detection and pose estimation - integrates several *ViTPose++-small* models [19] to detect objects (e.g., zebras and other quadrupeds) and estimate their body postures via keypoint localization. The framework demonstrated adaptability across species such as horses, motivating its exploration for livestock applications.

## 2.1. ViTPose++

*ViTPose* [20] represents a leading transformer-based pose-detection architecture in computer vision. Vision Transformers (ViTs) employ attention mechanisms to capture long-range spatial dependencies, advancing image classification, object detection, and pose estimation [21]. ViTPose supports both top-down and bottom-up pipelines. In [19], researchers extended this framework to develop *ViTPose++*, which enhances task specialization and generalization without increasing inference cost. For quadrupeds—whose limb and head geometries differ substantially from humans - ViTPose++ separates feed-forward networks into task-specific and task-agnostic branches, mitigating pose-structure conflicts. Consequently, ViTPose++ surpasses ViTPose on benchmark datasets such as AP-10K [22] and APT-36K [18], achieving higher accuracy and robustness in multi-species keypoint detection.

## 2.2. ZebraPose

ZebraPose builds upon ViTPose++-small and employs high-resolution synthetic imagery for keypoint detection on zebras and related animals. The synthetic dataset was generated using the GRADE framework [23] within *Unreal Engine* [24, 25], simulating 10 habitat types- including grasslands, forests, and ice plains. The resulting SC dataset comprised 18,000 annotated images, enriched with bounding boxes, segmentation masks, depth, and vertex information. Each virtual scene contained 250 randomized 3D zebra models rendered in 1080p using three unmanned aerial vehicle (UAV) camera viewpoints. Despite this scale, performance suffered when applied to real or close-up imagery, as most synthetic zebras appeared distant and small in frame. Rescaling failed to close this generalization gap, underscoring domain-transfer challenges between synthetic and natural images.

The initial SC dataset included bounding-box annotations only. A subsequent version added 27 anatomical keypoints (e.g., hooves, knees, thighs, tail base and tip, ears, eyes, nose, neck, and body joints). Owing to its synthetic design, camera-pose metadata enabled accurate projection of keypoints onto the 3D models. Only zebras with bounding boxes exceeding 30 pixels were annotated to ensure reliable keypoint resolution, and loss computations were restricted to annotated individuals. Figure 1 illustrates an example of synthetic zebra imagery from ZebraPose [16] showing bounding boxes and annotated keypoints.

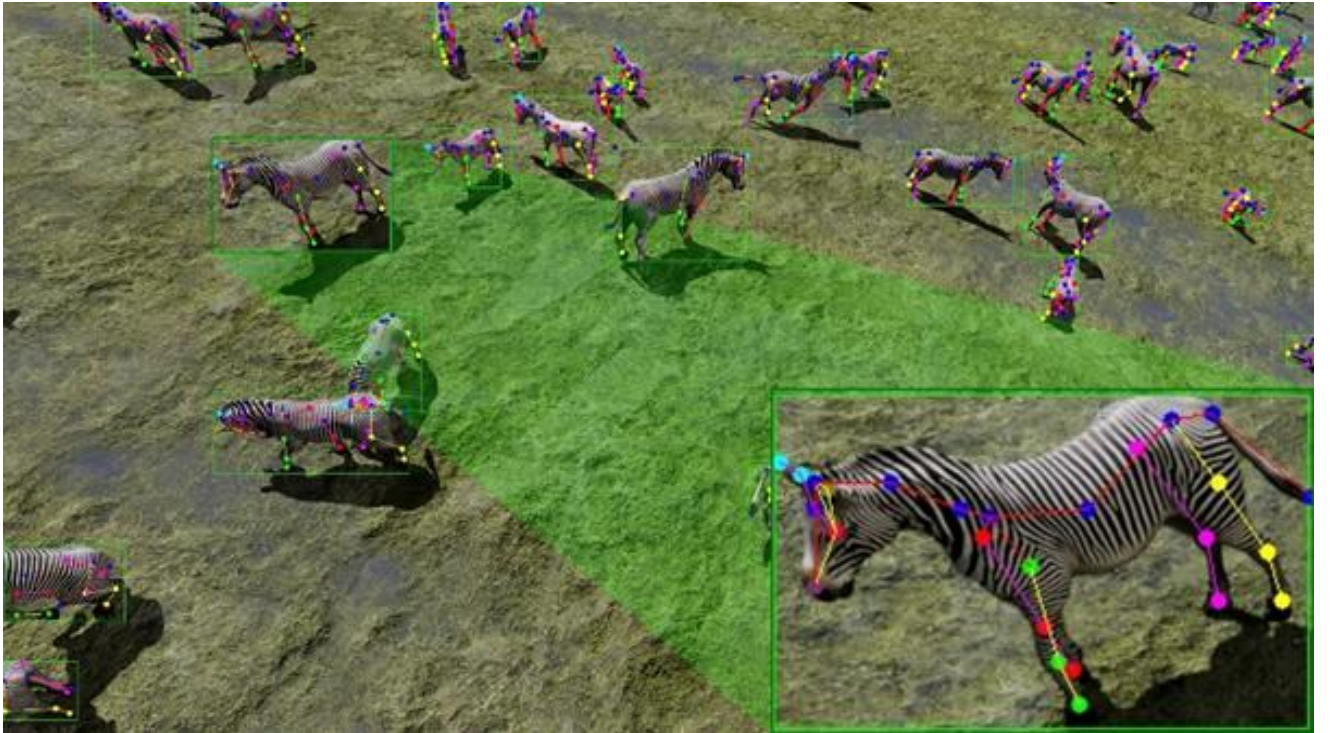

Figure 1. Annotated synthetic zebra images from ZebraPose [16], displaying bounding boxes and 27 keypoints generated using the GRADE-based virtual environment.

### 2.3. Performance Metrics

ZebraPose and ViTPose++ employ standard metrics for pose-estimation benchmarking.

#### 2.3.1. Percentage Correct Keypoints (PCK)

PCK quantifies the proportion of correctly localized keypoints within a defined threshold. A keypoint is deemed correct if the Euclidean distance between prediction and ground truth is below a specified fraction (e.g., 5 % or 10 %), expressed as PCK@0.05 and PCK@0.1 [26].

#### 2.3.2. Object Keypoint Similarity (OKS)

OKS measures spatial consistency by weighting each keypoint according to its scale and visibility [27]:

$$OKS = \frac{\sum_i exp\left(-\frac{d_i^2}{2 s^2 k_i^2}\right) \delta(v_i > 0)}{\sum_i \delta(v_i > 0)} \quad (1)$$

Here, $s$ is the bounding-box scale (square root of area), $d_i$ the distance between predicted and ground-truth keypoints, $k_i$ the keypoint-specific threshold, and $v_i$ the visibility flag. The threshold is determined as $k_i = 2\sigma_i$, where

$$\sigma_i = \sqrt{E[\frac{d_i^2}{s^2}]} \quad (2)$$

#### 2.3.3. Average Precision (AP)

AP integrates OKS across varying Intersection-over-Union (IoU) thresholds (0.5, 0.75, etc.) [28]. It represents the area under the precision–recall curve, with overall AP computed as the mean over 10 OKS thresholds (0.5–0.95 in 0.05 increments) [27].

#### 2.3.4. Average Recall (AR)

Analogous to AP, AR quantifies maximum recall across OKS thresholds [27], where recall is defined as [29]:

$$\text{Recall} = \frac{TP}{TP + FN} \quad (3)$$

## 2.4. ZebraPose Training and Evaluation

Multiple experiments on synthetic and mixed animal datasets validated ZebraPose performance using these metrics. A total of 23 ViTPose++-small models were trained with consistent hyperparameters: 210 epochs, decay steps at 170 and 200, Adam optimization, and a 0.0005 learning rate. Each model was evaluated across four zebra datasets, two multi-species datasets, and one horse dataset, and further tested using a Masked Autoencoder (MAE) pre-trained backbone to assess transfer benefits.

To assess cross-species transferability, the TigDog dataset [30] - containing tigers, dogs, and horses - was employed. The zebra-trained model was validated on the *horse subset* of TigDog, followed by experiments on mixed datasets:
(i) SC + TDH (complete TigDog), and (ii) SC + TDH99 (subset with 99 images). Table 1 summarizes PCK@0.05 and PCK@0.1 results for these configurations.

Table 1. Performance of ZebraPose trained on different datasets and validated on TigDog [30]. SC = synthetic zebra dataset [16]; TDH = TigDog dataset [30]; TDH99 = TigDog subset with 99 images.

| Train Set | Validation Set | PCK@0.05 | PCK@0.1 |
|---|---|---|---|
| SC | TDH | 0.276 | 0.450 |
| TDH | TDH | 0.965 | 0.968 |
| SC + TDH99 | TDH | 0.678 | 0.884 |
| SC + TDH | TDH | **0.987** | **0.989** |

The purely synthetic model exhibited the weakest generalization, reaffirming the synthetic-to-real performance gap. Incorporating even 99 real-animal images (SC + TDH99) markedly improved accuracy, illustrating the value of limited target-species data for fine-tuning. The mixed dataset (SC + TDH) achieved the highest PCK scores, outperforming both synthetic-only and real-only baselines—demonstrating the synergistic effect of hybrid data.

Building on these findings, we applied ZebraPose to dairy-cow imagery to evaluate whether similar cross-species adaptability could extend to livestock. The TigDog experiments suggested that modest quantities of annotated cow data might suffice to fine-tune a synthetic-trained model effectively-a promising direction for precision cattle monitoring in data-scarce agricultural environments.

## 3. Methodology

This study applies the ZebraPose framework to multi-animal datasets to evaluate its adaptability for dairy cow detection and pose estimation. The complete experimental workflow—from dataset preparation to model evaluation—is illustrated in Figure 2, which outlines the sequential stages of data collection, preprocessing, model training, and performance analysis.

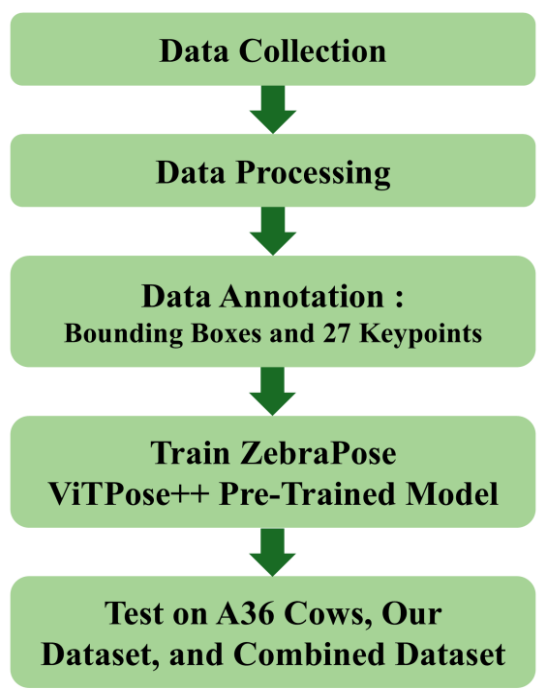

**Figure 2.** Workflow for cow detection and pose estimation using ZebraPose, showing the sequential stages of data acquisition, annotation, model training, and evaluation.

Two primary datasets and their combination were used to train and test ZebraPose: the benchmark APT-36K dataset [18] and a custom dairy-farm dataset collected in this study. The pre-trained ZebraPose model, originally trained on synthetic zebra data, was fine-tuned on each dataset separately and on their combination to examine transferability and generalization performance.

### 3.1. Datasets
This study utilized three datasets-A36 Cows, Our Dataset, and the Combined Dataset—to explore domain adaptation across synthetic, benchmark, and real barn environments.

### 3.1.1. A36 Cows
The original APT-36K [18] dataset includes images from 30 animal species with 17 annotated keypoints: left/right eyes, nose, neck, tail root, shoulders, elbows, front paws, hips, knees, and hind paws. Images were extracted from 2 400 online videos, with 15 frames per video, totaling 36 000 images. In [16], the 17 APT-36K keypoints were remapped to the 27-keypoint scheme used in ZebraPose, forming the derived A36 dataset.

For this research, only cow-labeled images were selected, producing the A36 Cows subset consisting of 960 images and 1 628 annotated cows. This subset provides controlled evaluation of model performance on an established benchmark dataset containing consistent keypoint annotations.

### 3.1.2. Our Dataset

The custom dairy-farm dataset was compiled from videos recorded in the dry-cow section of a commercial barn in Sussex, New Brunswick, Canada. Video acquisition occurred on May 7, 2025, between 10 a.m. and 6 p.m., using a GoPro HERO13 (UltraWide lens, 4 K resolution, 60 fps). Two videos—39 min 30 s and 35 min 14 s in duration—were selected for frame extraction. Frames were sampled every 12 seconds, ensuring sufficient pose and viewpoint diversity while minimizing redundancy. The barn housed five dry cows, typically with three to five visible per frame.

This process yielded 375 images featuring multiple cows in each scene. Annotations were independently performed by two annotators using CVAT [31] with the 27 ZebraPose keypoints, followed by cross-validation to ensure accuracy and consistency. The dataset captures realistic barn conditions including partial occlusions, varying illumination, and dynamic postures—factors often absent in benchmark datasets.

### 3.1.3. Combined Dataset

The Combined Dataset merged all images from both A36 Cows and Our Dataset, totaling 1 335 images and 3 584 annotated cows. This fusion introduced greater diversity in cow breed, posture, and environmental context, creating a balanced dataset for evaluating cross-domain robustness and data-fusion effects in agricultural pose estimation.

### 3.2. Model Training and Testing

All images underwent preprocessing in RoboFlow [32] to ensure uniformity and efficiency. Steps included auto-orientation to correct metadata rotations and resizing to 640 × 640 pixels. No data augmentation was applied to specifically examine transfer-learning performance under limited-data conditions.

Model training employed a pre-trained ZebraPose ViTPose++-small network originally trained on synthetic zebra images. The model was fine-tuned for dairy-cow pose estimation using identical hyperparameters across experiments: 210 epochs, learning rate = 0.0005, decay steps = 170 and 200, Adam optimizer, and Gaussian heatmaps. A Masked Autoencoder (MAE) pre-trained backbone was used to leverage robust feature initialization.

Three distinct models were trained:
- A36 Cows Model – trained on the benchmark subset;
- Our Dataset Model – trained on real barn imagery; and
- Combined Dataset Model – trained on the merged dataset.

For each configuration, ground-truth bounding boxes guided top-down keypoint detection. Although the default ZebraPose setup used 210 epochs, all three cow-based models showed continued improvement beyond this limit. Accordingly, training durations were extended (> 210 epochs) to identify optimal convergence while maintaining consistent optimization settings.

This systematic training-testing design enabled a controlled evaluation of transferability, domain adaptation, and generalization for cross-species pose estimation in realistic agricultural settings.

## 4. Experimental Setup and Results

As mentioned in the previous section, three ZebraPose ViTPose++-small models were used for the experiments on three datasets: 'A36 Cows', 'Our Dataset', and 'Combined Dataset' containing both. All datasets were divided into training, validation, and testing sets with a 70-20-10 distribution as shown in Table 2.

**Table 2.** Distribution of training, validation, and testing images across datasets (70–20–10 split).

| Dataset | Total Images | Training Set | Validation Set | Testing Set |
|---|---|---|---|---|
| A36 Cows | 960 | 672 | 192 | 96 |
| Our Dataset | 375 | 263 | 75 | 37 |
| Combined Dataset | 1335 | 935 | 267 | 133 |

All three models were implemented with a learning rate of 0.0005, decay steps at 170 and 200, Adam optimizer, Gaussian heatmaps, and the MAE pre-trained backbone. 'A36 Cows' model (i.e., model trained with 'A36 Cows' dataset), 'Our Dataset' model (i.e., model trained with our custom dataset), and 'Combined' model (i.e., model trained with combination of 'A36 Cows' dataset and 'Our Dataset') were executed for 320, 360, and 320 epochs, respectively. All other training parameters remained the same as the ZebraPose model. ViTPose automatically saves the model checkpoint file with the highest AP, therefore, AP, AR and PCK were used for testing the performances of the models.

### 4.1. 'A36 Cows' Model Performance

The ZebraPose model was trained and validated with the 'A36 Cows' dataset. Figure 3 shows training accuracy and training loss over 320 epochs indicating the training performance. As shown in Figure 3(a), after approximately epoch 200, the accuracy curve presented minimal variations, showing no significant changes and stabilizing around 80% training accuracy. Similarly, Figure 3(b) indicates that after approximately epoch 200, the model showed no further improvement. The training loss remained steady at around 0.0006 throughout the final 120 epochs.

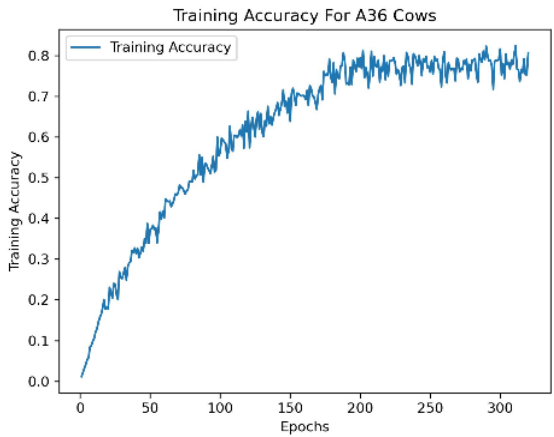
(a) Training Accuracy.

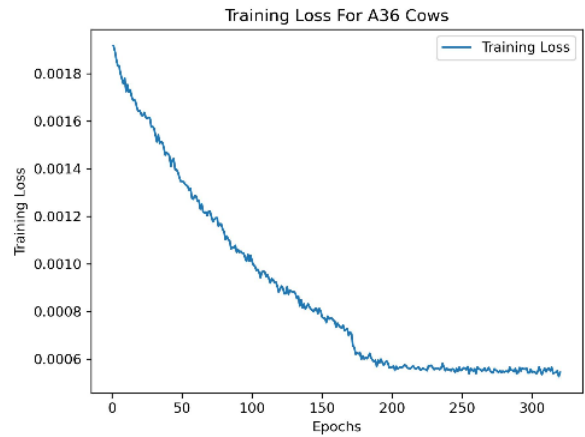
(b) Training Loss.

**Figure 3.** Training performance of the ZebraPose model on the A36 Cows dataset. (a) Training accuracy across 320 epochs showing early convergence near 80 %. (b) Training-loss curve stabilizing around 0.0006 after 200 epochs, indicating model saturation.

The model was validated after every 10 epochs by using the default settings from ZebraPose. Figure 4 shows the AP and AR of the model validation at 0.5 IoU and 0.75 IoU. Figure 4(a) shows the overall AP, AP at 0.5, and AP at 0.75 whereas Figure 4(b) shows the overall AR, AR at 0.5, and AR at 0.75. AR showed no noticeable variation after epoch 150, consistently remaining around 0.8 over the subsequent 17 validations. With the highest (quite close to 1) AP and AR at 0.5 and the overall AP and AR close to that, the model was able to detect most keypoints correctly with very few false positives.

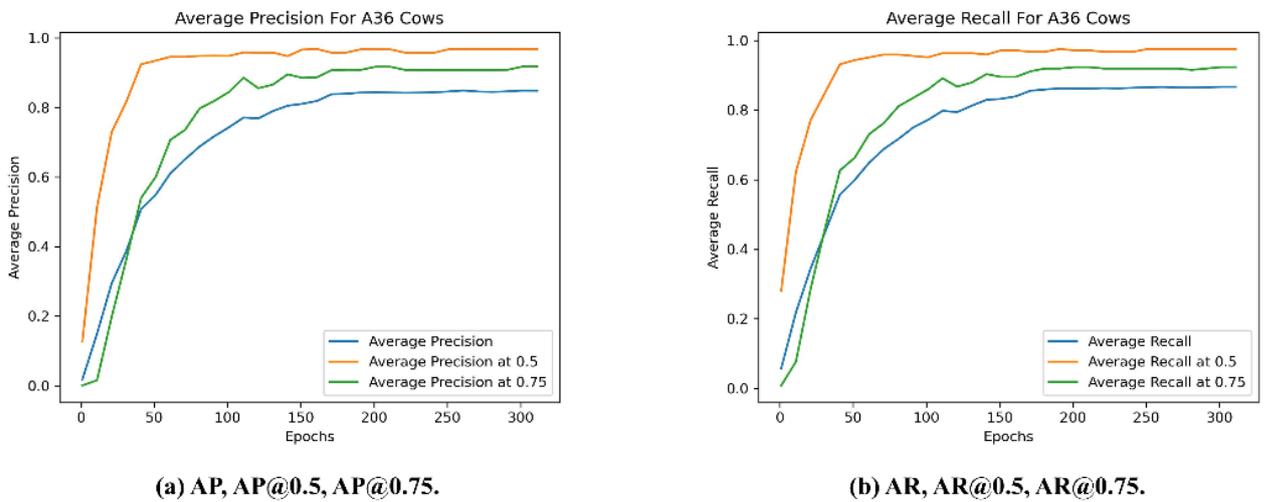

(a) AP, AP@0.5, AP@0.75.   (b) AR, AR@0.5, AR@0.75.

**Figure 4.** Validation metrics for the ZebraPose model trained on the A36 Cows dataset. (a) Average Precision (AP) trends at overall, 0.5 IoU, and 0.75 IoU thresholds. (b) Average Recall (AR) showing consistent high recall (~0.8) beyond 150 epochs.

Table 3 shows the results from the model trained and validated on 'A36 Cows', but tested on all datasets. Testing was performed on the testing sets (10% of the dataset) from each dataset after the 70-20-10 division. The AP, AR and PCK at 5%, 10%, and 20% were computed for comparing the performances.

**Table 3.** Performance metrics of the ZebraPose model trained on the A36 Cows dataset across all test sets. .

| Test Set | AP | AR | PCK@0.2 | PCK@0.1 | PCK@0.05 |
|---|---|---|---|---|---|
| A36 Cows | **0.849** | **0.867** | **0.943** | **0.883** | **0.779** |
| Our Dataset | 0.0000707 | 0.000748 | 0.146 | 0.0515 | 0.0127 |
| Combined | 0.384 | 0.336 | 0.352 | 0.267 | 0.216 |

As expected, the model achieved its highest performance on the 'A36 Cows' test set, likely due to the training and test sets originating from the same dataset sharing similar features. However, it failed to generalize to other datasets. When tested on the 'Combined Dataset', the model was only able to achieve a PCK@0.2 of 0.352, meaning it was able to correctly predict only 35.2% of the keypoints. Similarly, the model showed lower performance on 'Our Dataset' with a PCK@0.2 of 0.146. Although ZebraPose trained on 'A36 Cows' was able to perform well on a test set consisting of images from the same dataset as the training set, its inability to generalize to different datasets was a key limitation of this model.

## *4.2. 'Our Dataset' Model Performance*

The next model trained using the ZebraPose pre-trained model was using 'Our Dataset' for training and validation. Figure 5 shows the training accuracy and loss for the model. Figure 5(a) shows the training accuracy for the model and similar to the previous model, the accuracy showed no substantial increase beyond epoch 200 and remained at approximately 0.8 for the last 160 epochs. Although the training performance was very similar to the 'A36 Cows' model with about 80% accuracy, the performance fluctuations in the accuracy curve was minimized, showing more consistent performance. Similar characteristics were present in the loss curve in Figure 5(b) with approximately 0.0010 loss for the last 160 epochs showing slightly higher (about 4%) loss than 'A36 Cows' model.

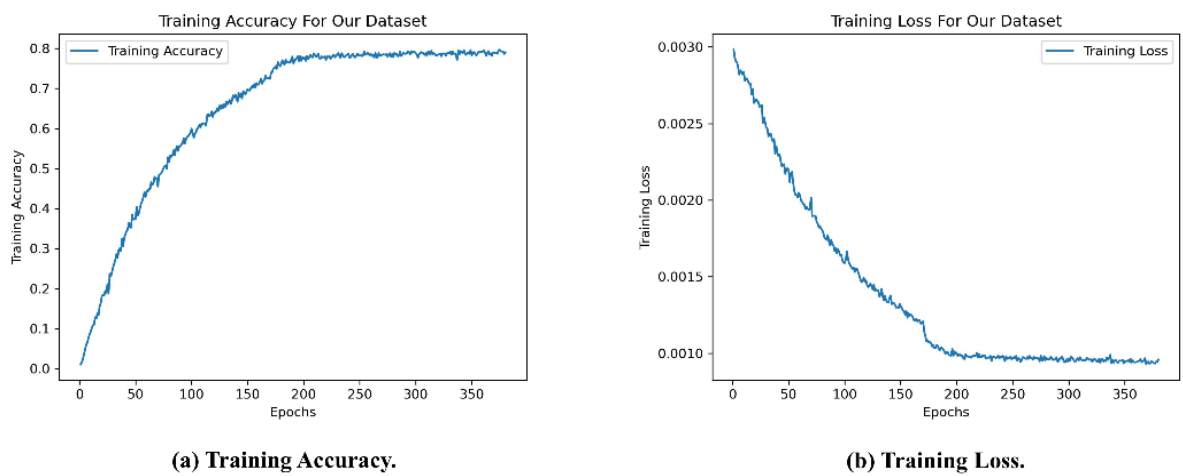

(a) Training Accuracy.        (b) Training Loss.

**Figure 5.** Training performance of the ZebraPose model on the custom barn dataset ("Our Dataset"). (a) Accuracy curve reaching ~80 % after 200 epochs with reduced variance.
(b) Loss curve flattening near 0.0010, indicating steady optimization.

The validation pipeline was invoked every 10 epochs by default for this model too. Figure 6 shows the AP and AR- overall AP, overall AR, AP at 0.5, AR at 0.5, AP at 0.75, and AR at 0.75. As shown in Figure 6(a), although the overall AP was steady after epoch 200, the highest AP was achieved at epoch 320 with a score close to 1. Interestingly, there was a large discrepancy between the AP at 0.5 and the overall average precision. This can indicate that the model was performing well at roughly locating the keypoints, but struggled with precise localizations of the keypoints. Similar patterns were visible for the AR in Figure 6(b) with very high AR at 0.5 but a large drop with a stricter IoU threshold.

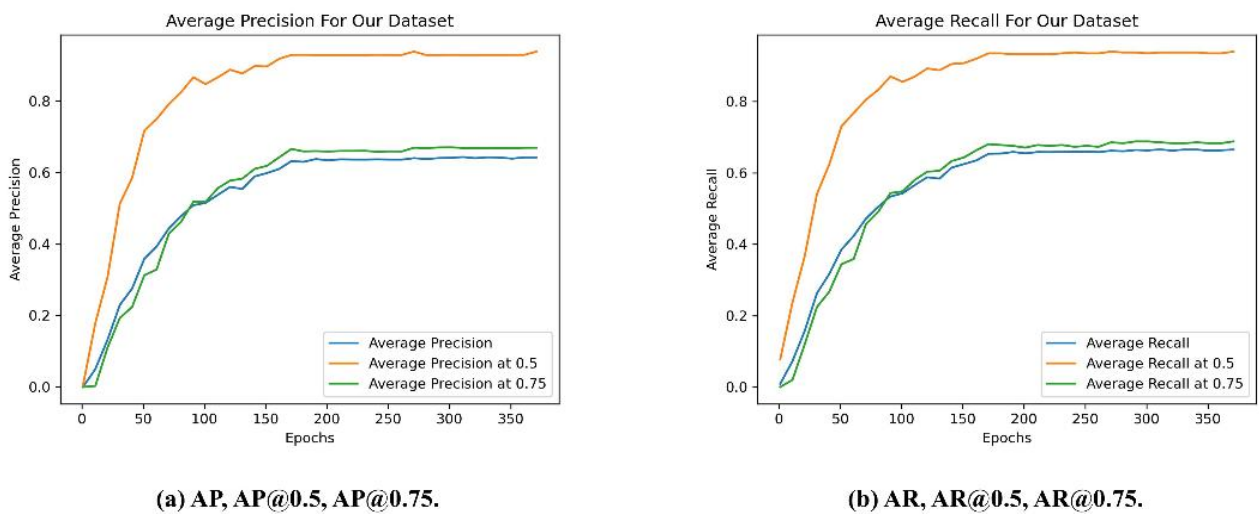

**(a)** AP, AP@0.5, AP@0.75.  **(b)** AR, AR@0.5, AR@0.75.

**Figure 6.** Validation performance of the ZebraPose model trained on the custom barn dataset. (a) Average Precision (AP) at overall, 0.5, and 0.75 IoU thresholds demonstrating strong localization but weaker fine-scale precision. (b) Average Recall (AR) showing high recall at 0.5 IoU and sharp decline at stricter thresholds.

Table 4 shows the results of the model trained and validated on 'Our Dataset' and tested on the testing sets from all three datasets. As shown in the table, this model failed to generalize to the 'A36 Cows' test set with an AP and AR of 0, indicating very poor performance. When testing on the combined dataset, we observed better results than 'A36 Cows' test set. With PCK@0.2 of 0.776, PCK@0.1 of 0.721, and PCK@0.05 of 0.640, the model was able to classify the majority of the keypoints correctly at all three thresholds. Consistent with previous results, the best performance was seen with the test set from 'Our Dataset'. These results show that while the model could perform well with images similar to its training data, it still struggled with generalizing to different environments, such as the 'A36 Cows' dataset.

**Table 4.** Performance metrics of the ZebraPose model trained on the custom barn dataset (Our Dataset).

| Test Set | AP | AR | PCK@0.2 | PCK@0.1 | PCK@0.05 |
|---|---|---|---|---|---|
| A36 Cows | 0 | 0 | 0.190 | 0.0838 | 0.0153 |
| Our Dataset | **0.644** | **0.666** | **0.931** | **0.864** | **0.715** |
| Combined | 0.470 | 0.483 | 0.776 | 0.721 | 0.640 |

### *4.3. 'Combined Dataset' Model Performance*

The final model was trained and validated on the 'Combined Dataset' containing data from both 'A36 Cows' and our custom dataset. Figure 7 shows the training accuracy and loss for this model. As seen in the graph in Figure 7(a), the accuracy remained stable after 200 epochs with the approximate accuracy of 0.8 for the last 120 epochs. Figure 7(b) shows the training loss for this model showing decrement for approximately the first 200 epochs and then remaining consistent with a score of 0.00075 for the final 120 epochs. The training accuracy and loss of the 'Combined' model were similar to the other two, but more closely resembled the second model trained with our custom dataset.

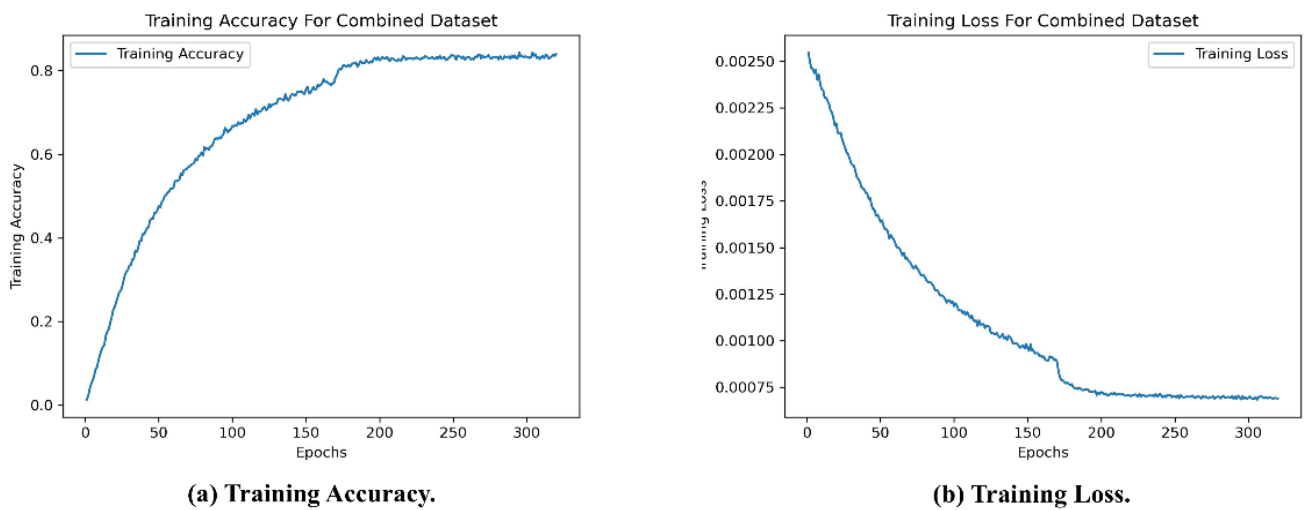

**Figure 7.** Training performance of the ZebraPose model on the Combined Dataset (A36 Cows + Our Dataset). (a) Training accuracy approaching 80 % and plateauing after 200 epochs. (b) Loss curve stabilizing near 0.00075, mirroring consistent learning behavior across datasets.

The validation AP and AR scores were computed at every 10 epochs as the other two models. Figure 8 shows the AP and AR curves. The AP curves in Figure 8(a) showing the overall AP, AP at 0.5, and AP at 0.75 and the AR curves in Figure 8(b) showing the overall AR, AR at 0.5, and AR at 0.75 showed a balance between the two previous models. The highest AP was achieved at epoch 190, and a large discrepancy existed between the AP at 0.5 and the overall AP. Interestingly, the gap was narrower than that of the second model, but wider than that of the first. This suggests that the model was able to correctly identify keypoints at lower thresholds but struggled at higher thresholds, indicating a lack of precise keypoint localization. Similar features were presented by the AR curve showing the model's difficulty with specifying the keypoints at their exact locations.

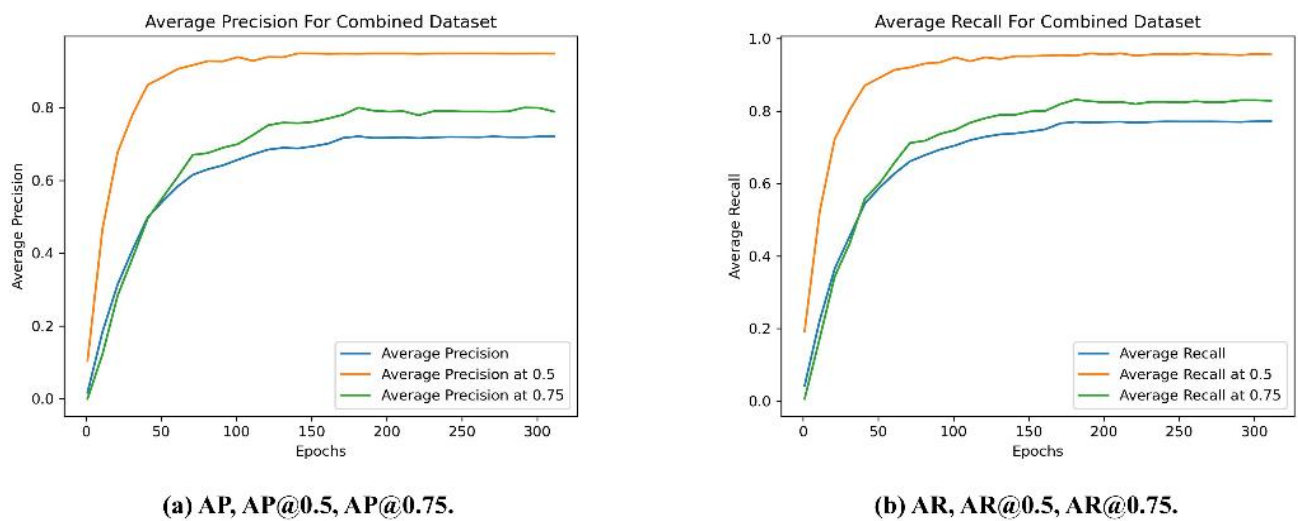

(a) AP, AP@0.5, AP@0.75.     (b) AR, AR@0.5, AR@0.75.

**Figure 8.** Validation metrics for the ZebraPose model trained on the Combined Dataset. (a) Average Precision (AP) curves across overall, 0.5, and 0.75 IoU thresholds showing balanced performance between domain sources. (b) Average Recall (AR) curves reflecting robust detection but reduced precision at higher thresholds.

Among the three models, the best results were achieved with the model trained on the 'Combined Dataset'. As shown in Table 5, the model performed best on our dataset's test set, with PCK scores of 0.977, 0.949, and 0.869 at thresholds of 0.2, 0.1, and 0.05, respectively. As this was the combined dataset, all the images in the test sets were similar to the training data. The AP, AR and PCK scores were consistent with the fact that all test datasets had similar data in training, hence the overall high performance among all three models.

**Table 5.** Evaluation results of the ZebraPose model trained on the Combined Dataset integrating A36 Cows and Our Dataset.

| Test Set | AP | AR | PCK@0.2 | PCK@0.1 | PCK@0.05 |
|---|---|---|---|---|---|
| A36 Cows | **0.860** | **0.879** | 0.930 | 0.897 | 0.812 |
| Our Dataset | 0.809 | 0.832 | **0.977** | **0.949** | **0.869** |
| Combined | 0.721 | 0.771 | 0.953 | 0.902 | 0.781 |

### *4.4. Overview of Model Performances*

Table 6 shows the highest and lowest performances for each dataset for easier comparison. Both 'A36 Cows' and 'Our Dataset' performed well when tested on data similar to their training data. However, they failed to achieve satisfactory results when tested on a different dataset. This implies that both models struggled with generalizing. The combined dataset performed well across all test sets, achieving the best results on our own dataset. However, since the combined dataset included images from both sources, the model may still face challenges with generalization.

**Table 6.** Summary of minimum and maximum AP, AR, and PCK@0.05 scores for all ZebraPose model configurations.

| Training Set | AP | | AR | | PCK@0.05 | |
|---|---|---|---|---|---|---|
| | Min | Max | Min | Max | Min | Max |
| A36 Cows | 0.0000707 | 0.849 | 0.000748 | 0.867 | 0.0127 | 0.779 |
| Our Dataset | 0 | 0.644 | 0 | 0.666 | 0.0153 | 0.715 |
| Combined Dataset | 0.721 | 0.860 | 0.771 | 0.879 | 0.781 | 0.869 |

### **5. Discussions**

The results included in the previous section represent some intriguing characteristics of the ZebraPose model on different training and testing datasets. The results with further analysis with the limitations of the models are discussed below.

### *5.1. Affect of Occlusions on OKS*

As our dataset included images from a real barn with real cow images, several cases of occlusions occurred due to the movements of the cows. During validation, the models trained on our dataset showed significantly better AP and AR scores at the 0.5 threshold, when compared to the 0.75 threshold and the overall score. This indicates that the model was able to predict keypoints that were slightly mislocalized. Therefore, there were many cases where the OKS was higher than 0.5 but lower than 0.75. It is important to note that this issue only occurred in models trained with our custom dataset. One key difference between 'Our Dataset' and 'A36 Cows' is the number of cows per image. Our dataset contained an average of 3 to 5 cows per image, whereas the 'A36 Cows' dataset contained an average of 2 cows per image. Therefore, our dataset suffered from more occlusions. This result supports the literatures as occlusions are known to decrease the performance of pose detection models [33]. Since lower OKS thresholds for AP and AR are known to be more tolerant of misplaced keypoints, mislocalized occluded keypoints may have an OKS value of greater than 0.5 but less than 0.75. Considering the occlusion possibility in real barn images, the models that trained on our custom dataset showed more realistic results for practical barn scenarios.

## 5.2. Performance on Test Sets

The generalizability concern was apparent from the model performance scores and graphs. The models trained only on 'A36 Cows' and only on 'Our Dataset' struggled to generalize to other datasets. While the models showed strong performances on their own test sets, the performance decreased significantly while tested on new data from another dataset. For the model trained with 'Our Dataset', this could be due to a lack of diversity in the dataset. 'Our Dataset' contained images from only the dry cows section of a barn, which housed five cows. While there were cows from other sections occasionally present in some of the dataset images, most of the dataset contained only five cows in the same environment. However, 'A36 Cows' contained images from various YouTube videos, therefore the images had a variety of backgrounds and cow breeds. Yet the model trained on 'A36 Cows' struggled to generalize to our dataset. Additionally, the majority of the images in 'A36 Cows' were taken from a lateral view, whereas our dataset included images taken from an aerial view. The differences in viewpoints could cause keypoints to be mislabeled. The lack of generalizability observed in the results of the models indicates the potential effects of different backgrounds, breeds, and camera angles in accurate keypoint localizations.

## 5.3. Visual Comparisons

Few keypoints visualization were generated for both the 'A36 Cows' and 'Our Dataset' images for easier visual comparisons. By comparing these with the ground truth keypoints, they provided valuable insights into the performance of the models. The predicted keypoints for one image from the 'A36 Cows' dataset using all three models, and the predicted keypoints for one image from 'Our Dataset' using all three models are presented for visual comparisons of the keypoints prediction generated by each model trained on 'A36 Cows' dataset, 'Our Dataset', and the 'Combined Dataset' respectively. These visualizations are meant to provide examples of how the model performed on one image from each dataset and therefore should not be taken as representatives of the entire dataset.

### 5.3.1. 'A36 Cows' Dataset Image

Figure 9 shows the predicted keypoints for the same image from 'A36 Cows' dataset using each of the three models. Figure 9(a) shows the ground truth keypoint annotations with 16 visible keypoints. Figure 9(b) shows 16 keypoints inferred by the model that was trained on 'A36 Cows' dataset (i.e., the same dataset this image is from). There were some limb alignment issues in the image and several keypoints were misplaced on the cow's back leg. In Figure 9(c), we can see how the model failed to identify any keypoints when trained only on 'Our Dataset'. Although this was not true for all images (as shown in Figure 10), this image still clearly represents the generalizability issue of the model. Finally, Figure 9(d) shows the performance of the model trained on the 'Combined Dataset'. The model predicted 18 keypoints but appeared to struggle to label the parts of the cow's leg correctly. Some parts of the limbs were swapped or not identified at all.

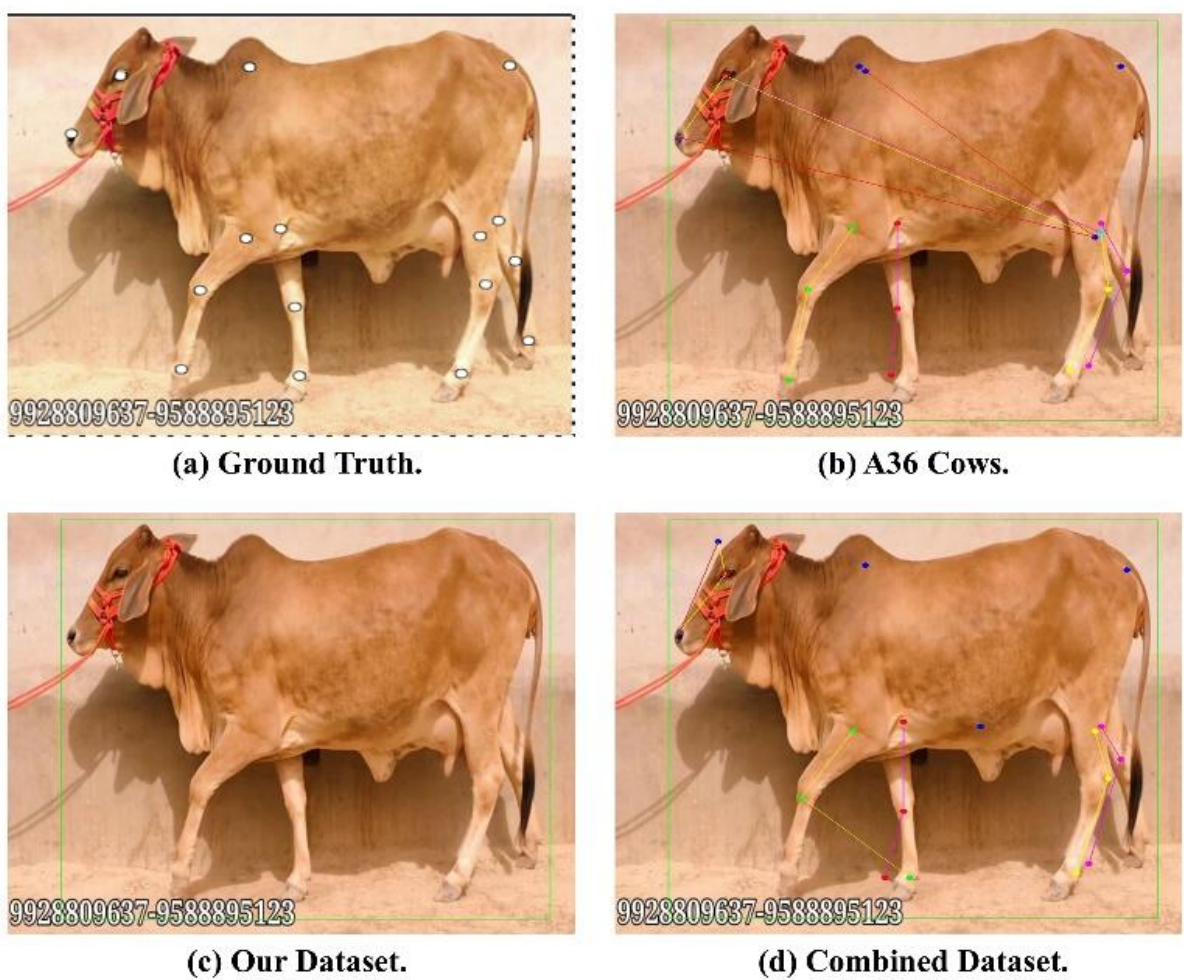

**Figure 9.** *Predicted keypoints versus ground-truth annotations for an A36 Cows image.* (a) Ground-truth keypoints (16 visible). (b) Predictions from the model trained on A36 Cows showing partial limb misalignment. (c) Model trained on Our Dataset failing to infer complete keypoints, revealing poor cross-domain generalization. (d) Model trained on the Combined Dataset identifying 18 keypoints with improved but imperfect limb association.

Figure 10 shows only two keypoints on another image from 'A36 Cows' using the model trained on 'Our Dataset'. While this performance is very poor, it shows that the model trained on only 'Our Dataset' was still able to make some predictions on the 'A36 Cows' images.

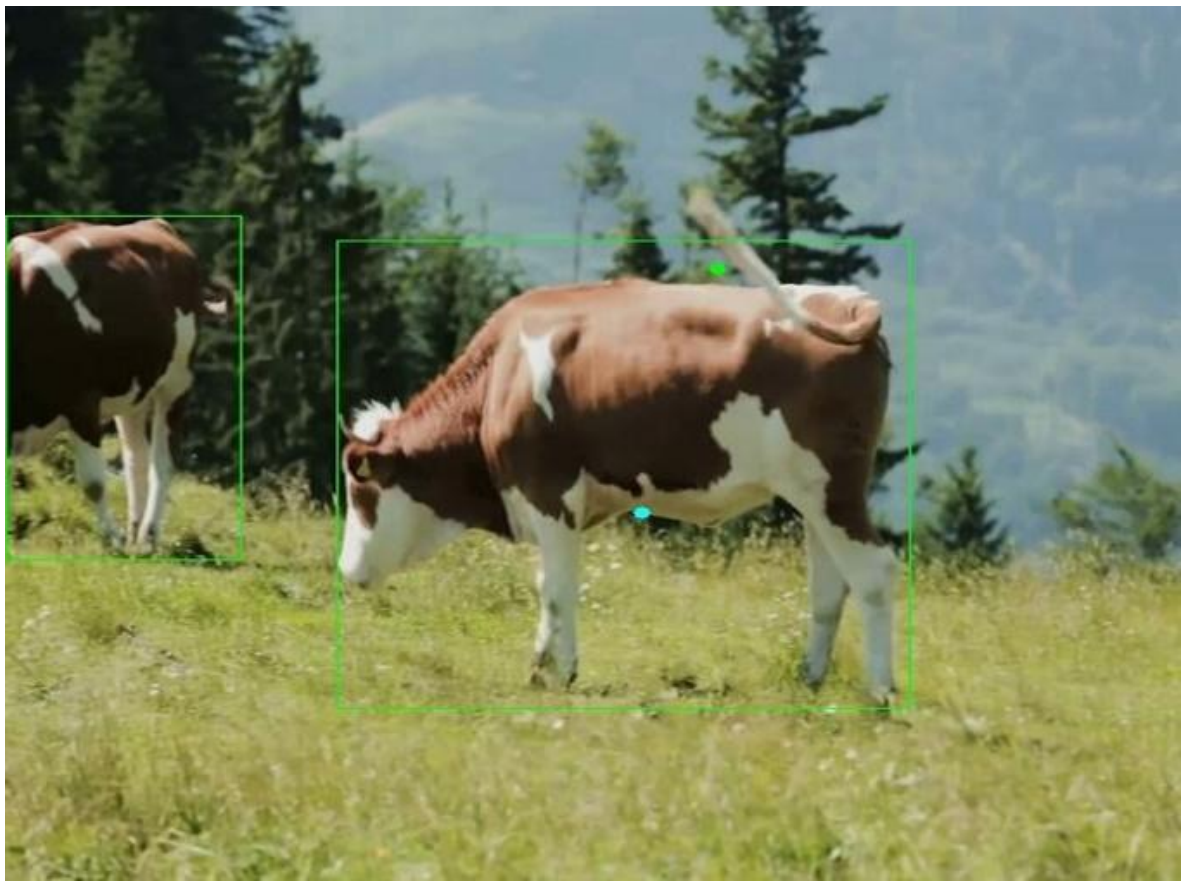

**Figure 10.** Example inference from the model trained on Our Dataset applied to an A36 Cows image. Only two keypoints are detected, illustrating severe domain-shift limitations when transferring from barn imagery to benchmark data.

None of the models were able to correctly identify all the keypoints from the ground truth annotations. However, all models except the one trained on 'Our Dataset' showed promising results. While they were unable to always identify the keypoints accurately, several keypoints, particularly the left eye and the nose were well detected. The model trained on o\'Our Dataset' failed to perform well on 'A36 Cows', possibly due to the lack of diversity in the training set. 'A36 Cows' is a much more diverse dataset that includes different breeds of cows in a variety of environments maintaining a rich data collection compared to ours.

### 5.3.2. 'Our Dataset' Image

Figure 11 shows the visual results of predicted keypoints on an image from 'Our Dataset' using all three models. As most images in 'Our Dataset' contained five cows in each frame, the bounding box of one cow was cropped for this visualization to ensure clear visualizations of the keypoints on one cow. Figure 11(a) shows the ground truth keypoints which consist of 21 visible keypoints and 6 occluded keypoints. Occluded keypoints are the ones within the frame but are hidden by another animal, object, or part of the cow's body and are shown as X's here. Figure 11(b) shows that the model trained on 'A36 Cows' struggled to predict any correct keypoints. There was only one keypoint identified within the cow's bounding box, however, it was predicted to be on the edge of the bounding box and not on the cow's body.

On the other hand, Figure 11(c) shows the predicted keypoints using the model trained on 'Our Dataset'. As expected, the predicted keypoints appeared to align closely with the ground truth keypoints. The performance was very similar on the 'Combined Dataset', as shown in Figure 11(d).

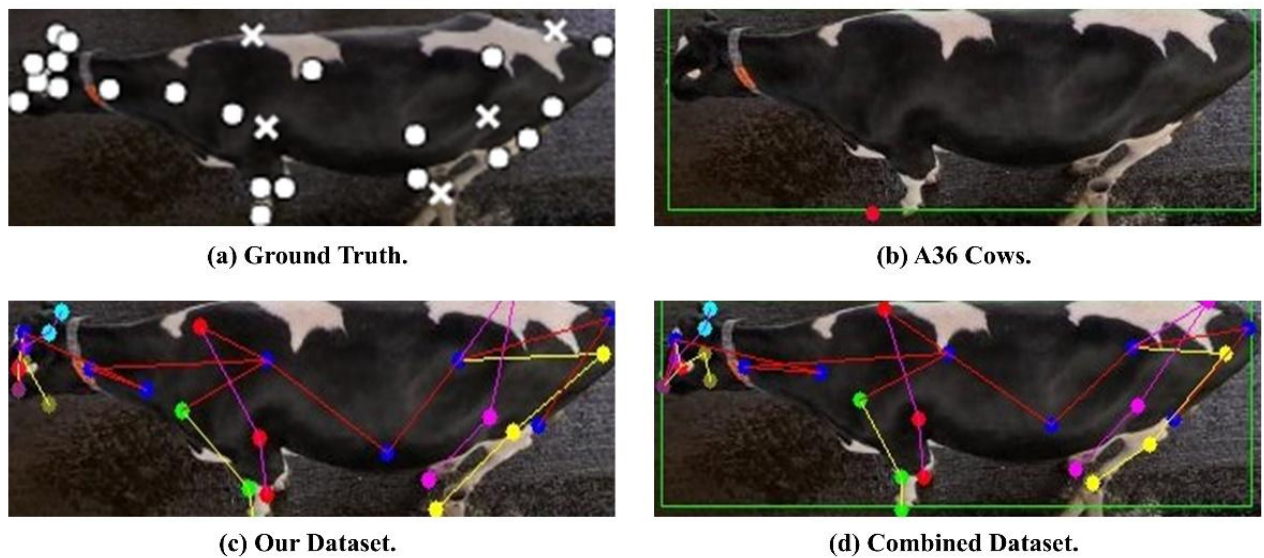

**Figure 11.** *Predicted and ground-truth keypoints for a sample image from Our Dataset.* (a) Ground-truth annotations showing 21 visible and 6 occluded keypoints. (b) Predictions from the A36 Cows-trained model misplacing most keypoints. (c) Predictions from the Our Dataset-trained model closely matching ground truth. (d) Predictions from the Combined Dataset model showing accurate localization comparable to in-domain training.

Although the sample visual examples are not the representatives of the entire datasets, they confirmed the model performances and limitations observed in the experimental results with visual representations. The examples showed how the predicted keypoints were consistent with the ground truth when the prediction model was trained on the same dataset the test image was from. Similarly, the images showed the same generalizability issue observed in the performance metrics. The model trained on one dataset performed poorly while predicting keypoints on images from the other datasets.

## 6. Conclusions

The agricultural AI revolution stands at a defining moment: the promise of computer vision–based livestock monitoring continues to inspire researchers and industry alike, yet our systematic evaluation of cross-species transfer learning exposes structural barriers that impede practical deployment in real farms. Using ZebraPose as a test case for dairy-cattle pose estimation, this study reveals a critical paradox in agricultural AI—models achieving high performance in controlled settings collapse when confronted with the diversity and unpredictability of real agricultural environments. Despite reaching AP, AR, and PCK@0.5 scores of 0.86, 0.87, and 0.869 under standardized conditions, complete generalization failure occurred across barns and herds, exposing the limitations of benchmark-centric evaluation that dominates current literature.

These findings reveal three pervasive misconceptions constraining progress. First, the synthetic-data fallacy assumes morphological similarity can substitute for ecological realism; yet zebra-trained models fail to translate across the environmental, behavioral, and contextual differences that define dairy systems. Second, the laboratory-performance illusion shows that metrics designed for stable imaging environments offer false assurance for deployment, as lighting fluctuations, seasonal changes, and occlusions inherent to barns invalidate benchmark reliability. Third, the dataset-scale delusion reflects the mistaken belief that small, homogeneous datasets - 375 images, five cows, one farm—can underpin scalable livestock AI, neglecting the heterogeneity of breeds, management styles, and regional climates that characterize global agriculture.

Addressing these misconceptions demands an agriculture-first paradigm in AI development. Future systems must be designed around the intrinsic variability of farming, prioritizing robustness over marginal metric gains. Model architectures should integrate temporal dynamics for behavioral context, environmental adaptation modules, and computational efficiency suitable for edge-based inference under resource constraints—embodying agricultural engineering principles rather than generic computer-vision assumptions. Evaluation standards must transition from laboratory benchmarks to on-farm validation protocols, where performance is measured by welfare insights, operational utility, and resilience to field conditions.

Meaningful advancement also requires diverse, globally representative datasets spanning multiple continents, breeds, and production systems to capture the true complexity of animal agriculture. Progress will depend on deep collaboration among computer scientists, agricultural engineers, animal behaviorists, and farmers, ensuring that innovation reflects practical realities rather than theoretical idealizations. As food-security pressures and sustainability imperatives intensify, the agricultural sector urgently needs reliable, explainable, and context-aware AI for livestock monitoring, welfare assessment, and behavioral analysis.

This study offers more than caution - it outlines a roadmap for transformation. By confronting current methodological blind spots, the agricultural AI community can redirect efforts toward systems genuinely grounded in farming realities. The path forward lies not in adapting general computer-vision models to agriculture, but in reimagining AI through an agricultural lens—one that values environmental complexity, operational robustness, and ethical stewardship. The field faces a pivotal choice: to persist with laboratory success that fails in barns, or to build AI systems capable of matching the intricacy, importance, and transformative potential of modern livestock farming.

**Acknowledgments:** The authors thank the Dairy Farmers of New Brunswick, Canada, for access to over 6 farms for data collection and for consultation and advice on the on-farm daily operations of dairy farming.

**Funding:** This work was kindly sponsored by the Natural Sciences and Engineering Research Council of Canada (RGPIN 2024-04450), Mitacs Canada, and the Department of NB Agriculture (NB2425-0025).

**Conflicts of Interest:** The authors declare no conflicts of interest.